\documentclass{article} 
\usepackage{colm2024_conference}

\usepackage{microtype}
\usepackage{xurl}

\usepackage{booktabs}
\definecolor{darkblue}{rgb}{0, 0, 0.5}
\hypersetup{colorlinks=true, citecolor=darkblue, linkcolor=darkblue, urlcolor=darkblue}

\usepackage[utf8]{inputenc} 
\usepackage{amsfonts}       
\usepackage{nicefrac}       
\usepackage{xcolor}         
\usepackage{amsmath}
\usepackage{tikz}
\usepackage{subcaption}
\usetikzlibrary{arrows.meta, positioning}
\usepackage{authblk} 

\usepackage[frozencache,cachedir=./minted-cache]{minted}
\usepackage{listings}

\newcommand{\ours}{\textit{Agent Lightning}}
\newcommand{\oursnorm}{Agent Lightning} 
\newcommand{\ourrlalg}{LightningRL}

\definecolor{mygray}{gray}{0.9}
\newcommand{\code}[1]{\colorbox{mygray}{\footnotesize{#1}}} 

\title{Agent Lightning: Train ANY AI Agents with \\ Reinforcement Learning}

\author{
\bf Xufang Luo\thanks{First authors.}, Yuge Zhang$^{*}$, Zhiyuan He$^{*}$, Zilong Wang, Siyun Zhao, Dongsheng Li, Luna K. Qiu, Yuqing Yang \\
Microsoft Research}

\begin{document}

\maketitle

\begin{abstract}

We present \ours{}, a flexible and extensible framework that enables Reinforcement Learning (RL)-based training of Large Language Models (LLMs) for any AI agent. 
Unlike existing methods that tightly couple RL training with agent or rely on sequence concatenation with masking, \ours{} achieves complete decoupling between agent execution and training, allowing seamless integration with existing agents developed via diverse ways (e.g., using frameworks like LangChain, OpenAI Agents SDK, AutoGen, and building from scratch) with almost ZERO code modifications. By formulating agent execution as Markov decision process, we define an unified data interface and propose a hierarchical RL algorithm, \ourrlalg{}, which contains a credit assignment module, allowing us to decompose trajectories generated by ANY agents into training transition. This enables RL to handle complex interaction logic, such as multi-agent scenarios and dynamic workflows. For the system design, we introduce a Training-Agent Disaggregation architecture, and brings agent observability frameworks into agent runtime, providing a standardized agent finetuning interface. Experiments across text-to-SQL, retrieval-augmented generation, and math tool-use tasks demonstrate stable, continuous improvements, showcasing the framework’s potential for real-world agent training and deployment.

\end{abstract}

\section{Introduction}

Recent advances in large language models (LLMs) have made AI agents effective for complex tasks like search, code generation, and tool use. These agents exhibit remarkable flexibility by leveraging LLMs' capabilities to adapt to diverse task requirements. While prompt engineering can help improve performance, LLMs still face significant limitations. They are prone to errors, especially when deployed in scenarios they weren't explicitly trained for--such as multi-turn coding workflows, private-domain datasets, or unfamiliar tools. Consequently, AI agents still struggle to reliably solve complex real-world tasks like end-to-end software development in products~\citep{liuLargeLanguageModelBased2024}. This limitation underscores the need to train or fine-tune models in agents to fully realize the potential of LLMs in these contexts~\citep{chenFireActLanguageAgent2023, jin2025search, songR1SearcherIncentivizingSearch2025, malooBitterLessonRethinking2025}.

Moreover, training models in real-world agents will likely become crucial for pushing the frontier of model capabilities. During agent execution, the system generates rich interaction data that captures the complexity of real-world problem-solving. These real-world experiences surpass traditional human-curated datasets in both scale and diversity, making it key for future LLM training~\citep{silver2025welcome,yaoSecondHalf,kimik2}. Consequently, leveraging this data for fine-tuning not only refines an agent's specialized skills but also fosters the development of more versatile LLMs suited for dynamic, interactive environments.

\begin{figure}[tbp]
    \centering
    \includegraphics[width=0.8\textwidth]{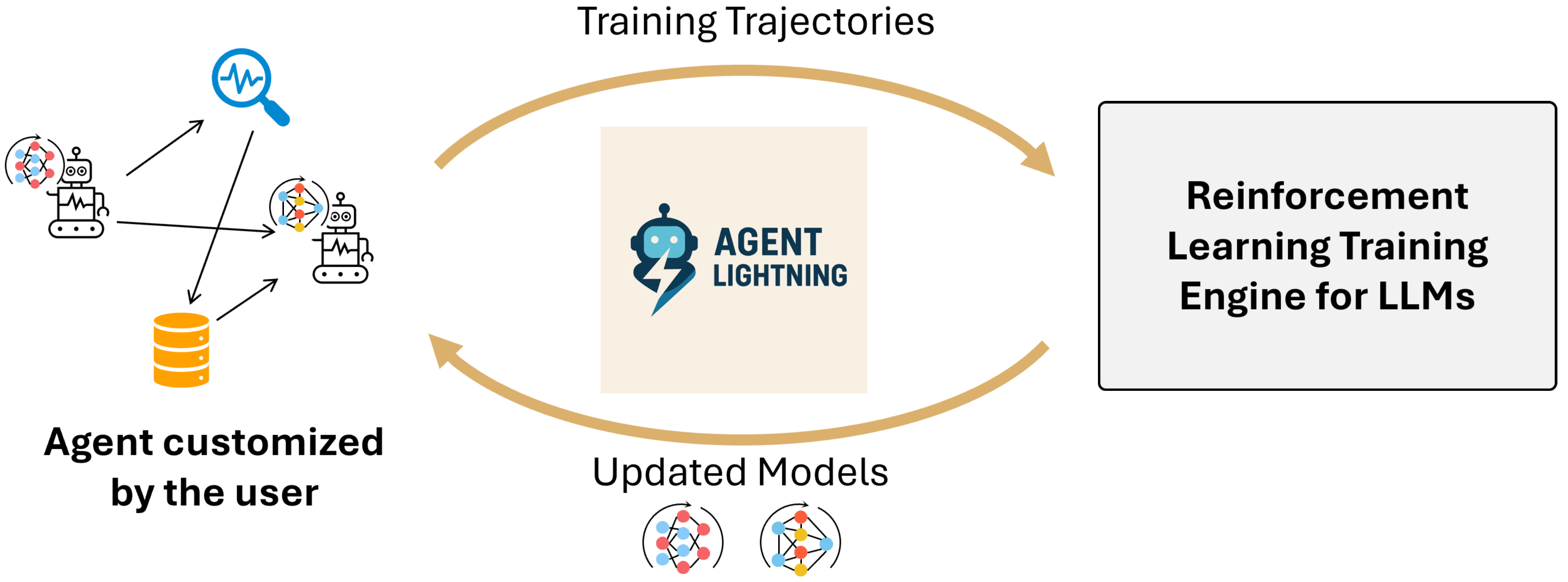}
    \caption{Overview of \ours{}, a flexible and extensible framework that enables reinforcement learning of LLMs for ANY AI agents.}
\label{fig:intro_overview}
\end{figure}

Reinforcement learning (RL), which has driven recent advances in reasoning models such as DeepSeek-R1~\citep{guo2025deepseek} and Kimi k1.5~\citep{team2025kimi}, offers a powerful paradigm for optimizing LLMs in agentic scenarios. While supervised learning requires detailed step-by-step annotations--which are scarce and costly for complex interactive tasks--RL relies on outcome-based reward signals. This eliminates the need for task-specific curated data and allows agents to learn desirable behaviors directly from environment feedback across diverse tasks. Moreover, the trial-and-error nature of RL closely mirrors how humans acquire problem-solving skills, enabling models to learn action policies grounded in deployment contexts. This capability opens up the potential for transforming LLM-generated text tokens into real-world actions, making RL a natural fit for training models in agent-based systems.

However, extending the RL paradigm to agents introduces substantial challenges in both algorithm design and system implementation.
Existing RL methods and frameworks for LLMs are primarily tailored for static, single-call tasks, such as preference alignment or mathematical reasoning. In contrast, agents exhibit both complexity and diversity beyond this setting. The complexity lies in the fact that their execution often involves multiple LLM invocations with distinct prompts and responses, as well as interactions with external tools, APIs or environments. The diversity stems from the need to design different agents tailored to the requirements of various applications. These challenges hinder the application of RL for large-scale LLM tuning in agents.

We introduce \ours{}, a flexible and extensible framework that enables RL-based training of LLMs for ANY AI agent. As shown in Figure~\ref{fig:intro_overview}, \ours{} achieves \textit{complete decoupling between agent execution and RL training}, empowering developers to train existing agents with almost ZERO code change.

This decoupling is grounded in formulating agent execution as a Markov decision process (MDP), where the \textit{state} represents the current snapshot of agent execution, comprising variable values that sufficiently describe the execution status. The \textit{action} corresponds to the output generated by the policy LLM, which is then used to update the state. Building on this formulation, we propose a unified data interface for RL training where agent trajectories are structured as sequences of transitions. Each transition contains the current state (LLM input), action (LLM output), and reward.
This unified data interface abstracts away the underlying orchestration logic and agent framework details, making it applicable to ANY agent.
Furthermore, to optimize the policy LLM using collected transitions, we introduce \ourrlalg{}, a hierarchical RL algorithm specifically designed for agent training. After assigning credit to each transition, \ourrlalg{} updates the policy in a manner that is fully compatible with existing single-turn RL methods for LLMs.
Such modeling and algorithm offers many benefits, including enabling highly flexible context construction, allowing selective optimization of multiple agents simultaneously, and relieving the issue of accumulative context leading to excessively long sequences.

Building on the concrete formulation described above, \ours{} introduces a Training-Agent Disaggregation (TA Disaggregation) architecture to establish a standardized training service applicable to any agent. The framework consists of a Lightning Server and Lightning Client. The Lightning Server functions as the controller of the RL training system, managing the training process and exposing an OpenAI-like API of the updated model to the client. The Lightning Client comprises two functional components: one handles communication with the server for data transmission and reception; the other runs the agent and performs data collection, serving as the agent runtime.

Benefiting from the unified data interface, the agent runtime enables the leverage of the comprehensive observability frameworks like OpenTelemetry~\citep{OpenTelemetry} into training process for trajectory collection. This connects observability infrastructure with RL training, enabling optimization techniques to use rich monitoring data, and creating a foundation that enhances scalability and flexibility. 
The agent runtime can also facilitate the algorithms to mitigate the sparsity of rewards, by the \textit{Automatic Intermediate Rewarding (AIR)} mechanism. AIR enables the assignment of intermediate rewards to transitions based on system monitoring signals (such as tool call return statuses). Developers can easily customize this mechanism, effectively mitigating the sparse reward problem in agents, allowing for more effective training.

We demonstrate \ours{}'s effectiveness through various tasks, including a text-to-SQL agent implemented with LangChain, a retrieve-augmented generation agent built with OpenAI Agents SDK, and a math question answering agent with tool usage developed via AutoGen. The reward curves of these experiments demonstrate that \ours{} enables continuous and stable performance enhancement across different agent scenarios, showing great potential for solving real-world tasks.

In summary, \ours{} advances agentic training in the following aspects.
\begin{itemize}
    \item \ours{} is the first framework to achieve a full decoupling between agents and RL training. This decoupling enables \ours{} to be seamlessly applied to ANY AI agent regardless of implementation approach, with almost ZERO code modifications. By aligning training with the agent's execution logic, it directly enhances agent performance in real-world applications. This empowers developers to go beyond static, pre-trained models and unlock the full potential of adaptive, learning-based agents. It also acts as an unified data interface, collecting diverse agent-generated data to improve model capabilities. To achieve this decoupling, \ours{} introduces specialized designs in both algorithm and infrastructure.
    \item For the algorithm, \ours{} builds on the MDP formulation of agents and introduces a unified data interface. This interface abstracts away the complexity of various agent execution logic, allowing data collected during agent execution to be directly transformed into training trajectories. Additionally, Agent Lightning employs a hierarchical RL framework with a credit assignment module that allocates trajectory-level returns to responses generated by each call. This design integrates seamlessly with existing single-turn RL algorithms, enabling efficient and effective training.
    \item For the system, \ours{} introduces a Training-Agent Disaggregation architecture that achieves a clean separation between RL training and agent execution. This architecture is realized through the Lightning Server and Lightning Client, which together provide a standardized model training service applicable to any agent. The Lightning Client functions as the agent runtime, transparently managing agent execution and collecting trajectories without requiring code modifications. This design enables the reuse of the observability infrastructure in training scenarios, ensuring extensibility, scalability, and seamless integration with diverse agent frameworks.

\end{itemize}

\section{Modern AI Agents}

Precisely defining AI agents is challenging due to their diversity and rapid evolution. While several works have attempted to formalize and categorize AI agents~\citep{significantgravitasAutoGPT2023, shenHuggingGPTSolvingAI2023, BuildingEffectiveAI2024,HowWeBuilt2025, KimiresearcherEndtoendRL2025, xuComprehensiveSurveyDeep2025}, our goal is not to introduce a new definition. Instead, we provide an abstracted formulation that serves as a common understanding of AI agents: \textit{an AI agent is a software system that incorporates one or more LLM calls within its execution}. This broad definition encompasses a wide spectrum of existing AI agents, ranging from workflow-based systems that follow predefined code paths~\citep{BuildingEffectiveAI2024} to advanced multi-agents capable of dynamic planning, reasoning, and long-horizon operation~\citep{HowWeBuilt2025, KimiresearcherEndtoendRL2025, xuComprehensiveSurveyDeep2025}.

In this section, we present a high-level introduction of modern AI agents, including common components used in agent construction, typical orchestration of these components, and agent frameworks for easily developing agents.

\subsection{Component}

In general, the building components for agents can be divided into two types.
\begin{itemize}
    \item \textbf{LLMs}, or more generally, foundational models (FMs), serve as the core reasoning and generation engines within agents. Every LLM generation is a stateless mapping from an input (prompt) to an output (response).
    Due to their substantial computational requirements, LLMs typically run on high-performance servers or cloud platforms, with agents accessing them via API endpoints -- either through commercial providers (e.g., OpenAI, Google) or self-hosted solutions (e.g., vLLM~\citep{vllm}, SGLang~\citep{sglang}). 
    An agent may leverage multiple LLMs, each identified by a unique endpoint URL, to perform different tasks or achieve diverse functionalities. We define the set of LLMs used by an agent as $\mathcal{M} = \{ M_i \}_{i=1}^m$, where $m$ is the number of LLMs in the agent. Each $M_i$ can be a different model or version, and we can link the model name to its parameters $\theta_i$ for clarity.
    \item \textbf{Tools} are functionalities that the agent invokes to perform specific tasks, such as retrieving information from a database, executing code, or interacting with APIs. Tools can be stateless or stateful, depending on whether they maintain context across invocations. Besides, tools can be implemented in various ways, including external APIs, local programs, or libraries. In recent trend, tools often follow the MCP (Model Context Protocol)~\citep{mcp} or other agent communication protocols to standardize interactions between agents and tools, enabling seamless integration and interoperability. Let $\mathcal{T} = \{ T_j \}_{j=1}^t$ denote the set of tools available to the agent, where $t$ is the number of tools. 
\end{itemize}

\subsection{Orchestration}
\label{sec:2_orchestration}

Using the component set $\mathcal{M} \cup \mathcal{T}$ as functional building blocks, orchestration -- including the execution flow, dependencies, and order among components -- is typically customized by the user based on task requirements.
A detailed example of an agent and its execution flow is provided in Section~\ref{sec:example_agent_execution}.

Importantly, this orchestration is often neither fixed nor deterministic. Modern AI agents exhibit dynamic behaviors where LLMs may determine subsequent actions or select tools based on the evolving context. For instance, by letting the LLM to decide to either refine the query or directly answer the question, the above RAG agent can have variable number of interaction turns with the database. Another example is the game-playing agent. The iterative process of generating actions and receiving feedback from the environment is represented as a sequence of LLM calls, where each call corresponds to a decision-making step in the game. By unrolling this iterative process, we can capture the dynamic behavior of the agent, where the LLM may adapt its strategy based on the current game state and previous actions.

While this dynamic interaction 
captures rich task-specific behaviors that could enhance optimization, it also poses significant challenges for data modeling and downstream learning, making it impractical and unscalable to implement different agents with current RL training frameworks.

\subsection{Implementation}

Developers have multiple approaches for creating AI agents. Some may choose to build agents entirely from scratch for complete control and customization. Some other developers may leverage established agent development frameworks that provide infrastructure and accelerate development. These frameworks include orchestration platforms like LangChain~\citep{LangChain}, OpenAI Agents SDK~\citep{oaiagentsdk}, and AutoGen~\citep{AutoGen}. 

Agent frameworks facilitate development by providing modular components (LLMs and Tools) and flexible mechanisms for dynamic orchestration, enabling users to define workflows that adapt to both task requirements and the agent's runtime state. However, regardless of whether agents are built from scratch or using frameworks, they typically lack mechanisms for automatic self-improvement and do not involve agent optimization. In real-world scenarios, especially when dealing with private data, the performance of agents often falls short of requirements, necessitating an automated optimization.

 \section{Agent Lightning}

We now illustrate the proposed \ours{} framework. We begin by defining the unified data interface in Section~\ref{sec:method_udi}, which specifies what data is collected during agent execution, agnostic to the agent design. Next, in Section~\ref{sec:method_mdpa}, we show the Markov decision process in agents, lying down the foundation for reinforcement learning (RL). Then, in Section~\ref{sec:method_hrl}, we present our hierarchical RL algorithm that leverages the collected data in agents to update the LLM. Finally, we describe our tailored infrastructure with several elegant design elements that significantly streamline the agent training process in Section~\ref{sec:method_infra}\footnote{Note that to ensure clear and precise referencing, in our paper, execution refers to the collected data during one agent run. Trajectory/episode emphasizes extracted data used for RL training.}.

\subsection{Unified Data Interface}
\label{sec:method_udi}

Data-driven agent optimization fundamentally depends on the data generated during agent execution. Therefore, in \ours{}, we define a unified data interface that specifies how this data feeds into RL training algorithms. This interface is highly general and can be applied to data collected from any AI agent.

AI agent is a special kind of software, and like all software execution, can be represented as a directed acyclic graph (DAG)~\citep{alfred2007compilers,abadi2016tensorflow}. In this graph, each node corresponds to a component invocation (either an LLM or a tool) from the set $\mathcal{M} \cup \mathcal{T}$, while each edge represents a dependency or control flow between components. However, parsing a complete graph from disorganized agent execution traces is non-trivial and, as we have discovered, unnecessary for training purposes. By leveraging the concept in Markov Decision Processes (MDPs), we only need to identify the current \textit{state} and the key factors influencing state transitions (i.e., \textit{calls}) to perform RL optimization.

\subsubsection{State and Call}
\label{sec:state_call}

We define the $\texttt{state}$ as a snapshot of the agent execution, which may include information such as the program counter, variable values, call stack, resource context, and more. These can be abstracted as a set of variables that evolve over time as the agent executes. The same task $x$ can be executed multiple times, producing different execution traces due to the dynamic behavior of agents. For the $k$-th execution of $x$, $\texttt{state}$ at timestep $t$ containing $V$ variables is represented as:
\begin{equation}
\label{eq:state}
\texttt{state}_t(x,k) = \left\{ \texttt{variable}_i^{x,k,t} \right\}_{i=1}^{V}.
\end{equation}

\begin{figure}[tbp]
    \centering
    \includegraphics[width=1.0\textwidth]{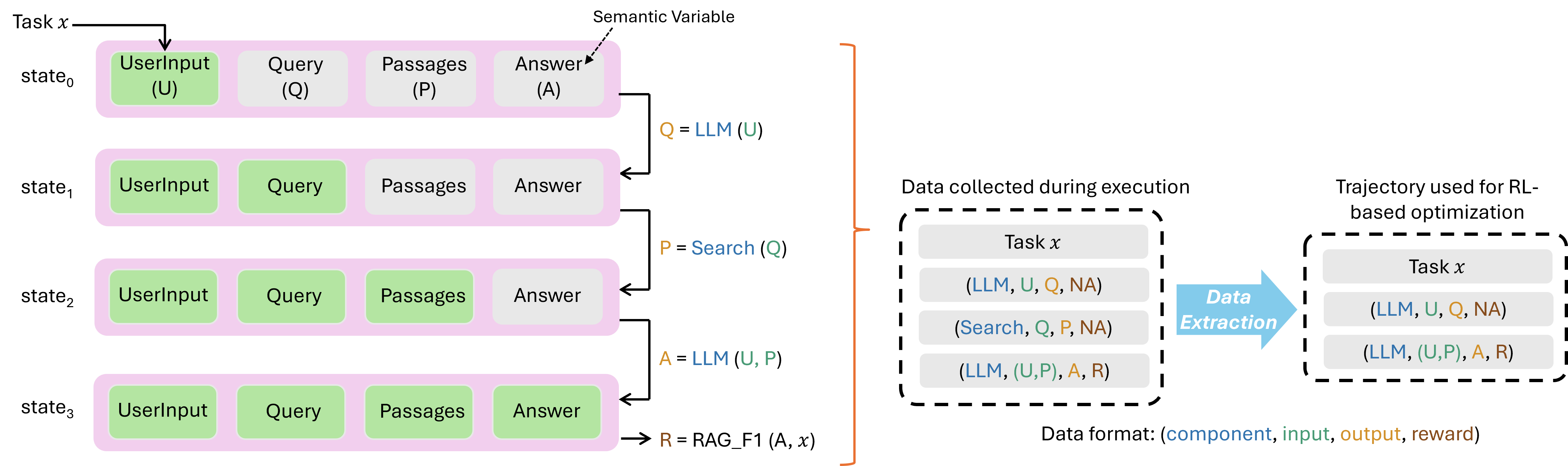}
    \caption{Illustration of the unified data interface in \ours{}. The left panel depicts the agent execution flow, where each state transition is represented by the update of semantic variables (green rectangles denote variables with valid values; gray rectangles indicate variables not yet assigned in the current state). The right panel presents the corresponding trajectory collected throughout the agent's execution, demonstrating how the unified data interface systematically captures all relevant transitions for RL-based optimization.}
    \label{fig:data_interface}
\end{figure}

The state changes continuously during agent execution.  For example, the counter in a for loop increments, and simple string manipulations modify intermediate variables. However, these intermediate operations and auxiliary code are not essential for agent optimization. 
Here, we primarily focus on key variables -- those used or modified by components and represent critical semantics (such as program intents) -- which we call  \textit{Semantic Variable}~\citep{lin2024parrot}. A concrete example on states and semantic variables is provided in Section~\ref{sec:example_agent_execution} and Figure~\ref{fig:data_interface}. 

Changes to Semantic Variables occur through \textit{Component Invocation} of agents.
Assume that the $k$-th execution of $x$ consists of $N$ component invocations/calls:
\begin{equation}
\label{eq:agentexecution}
\texttt{execution}(x,k) = \left\{ \texttt{call}_i^{x,k} \right\}_{i=1}^{N},
\end{equation}
where the $i$-th component invocation is defined as a $\texttt{call}$:
\begin{equation}
\label{eq:call}
\texttt{call}_i^{x,k} = (\texttt{meta}_i^{x,k}, \texttt{input}_i^{x,k}, \texttt{output}_i^{x,k}), \quad\text{with}~\texttt{output}_i^{x,k} = C_i(\texttt{input}_i^{x,k}).
\end{equation}
Here, $C_i \in \mathcal{M} \cup \mathcal{T}$ denotes the component being invoked (either an LLM or a tool) in this call, $\texttt{input}_i^{x,k}$ and $\texttt{output}_i^{x,k}$ represent the input and output data. $\texttt{meta}_i^{x,k}$ contains relevant meta information about the invocation, such as $C_i$'s name, type, version, API endpoint, as well as runtime parameters such as sampling strategy and temperature for LLM calls.

As the $\texttt{state}$ represents a snapshot of the agent, all inputs and outputs are semantic variables at specific timesteps:
\begin{equation}
\label{eq:inout}
\texttt{input}_i^{x,k}\in\texttt{state}_{t_1}(x,k), \quad \texttt{output}_i^{x,k}\in \texttt{state}_{t_2}(x,k).
\end{equation}
Not all information in the $\texttt{state}$ is necessary or visible to the component being called. For instance, in Figure~\ref{fig:data_interface}, although the $\texttt{state}$ contains 4 semantic variables, the LLM can only see the user input when writing the first query and the search tool only takes the generated query as input.

\subsubsection{Reward and Dataset}

To enable learning from collected data during agent execution, we introduce reward signals to measure task completion quality. Each agent execution is augmented with scalar rewards $\{r_1, \dots, r_N\}$, where each $r_i \in \mathbb{R}$ corresponds to the quality of $i$-th invocation. An execution with its reward signals is represented as:
\begin{equation}
\label{eq:execution_r}
\texttt{execution}^R(x,k) = \left\{ (\texttt{call}_i^{x,k},r_i^{x,k}) \right\}_{i=1}^{N}.
\end{equation}

Reward signals can be provided at either intermediate steps or at the conclusion of an agent's execution, depending on the task and agent design. Intermediate rewards $r_i$ ($i < N$), such as indicators of successful tool invocation or partial task completion, provide more granular feedback and guide the agent's learning process. The final reward $r_N$ typically assesses the agent's overall success in accomplishing the task. In many real-world scenarios, only a terminal reward $r_N$ is available to assess the outcome of the entire execution. This represents a special case of the general formulation where intermediate rewards $r_1, \dots, r_{N-1}$ are absent and learning depends solely on the final result.

A dataset for training or evaluation consists of a task set $\mathcal{X}=\{x_1,\dots,x_{|\mathcal{X}|}\}$ and a reward function, where $|\cdot|$ denotes the set size. Tasks can be executed multiple times, with executions annotated according to the reward function.

\subsubsection{An Illustrative Example}
\label{sec:example_agent_execution}
To concretely illustrate agent execution and the unified data interface, we now consider a typical retrieve-augmented generation (RAG) agent, shown in Figure~\ref{fig:data_interface}. This agent is designed to answer user queries by first retrieving relevant documents and then generating a response grounded in those documents. The agent comprises two main components: the generative policy model ($\mathcal{M} = \{\mathrm{LLM}\}$) and a search tool ($\mathcal{T} = \{\mathrm{Search}\}$).

The execution flow proceeds as follows:
\begin{enumerate}
    \item The user submits a specific task ($x$) with the question (\texttt{UserInput}).
    \item The agent invokes the LLM to generate a search query (\texttt{Query}) conditioned on the \texttt{UserInput}.
    \item The Search tool retrieves relevant passages (\texttt{Passages}) using the generated \texttt{Query}.
    \item The agent calls the LLM again to produce a final answer (\texttt{Answer}), leveraging both the retrieved \texttt{Passages} and the original \texttt{UserInput}.
\end{enumerate}

Upon completion, a reward function (e.g., \texttt{RAG\_F1}) evaluates the quality of the generated \texttt{Answer}.
With our formulation, each \texttt{state} encapsulates the current values of semantic variables (green rectangles in Figure~\ref{fig:data_interface} denote variables with valid values, while gray rectangles indicate variables without values in the current state). As the agent executes, invoking a component (LLM or tool) updates the relevant semantic variable, resulting in a change to the next state. The agent's execution for a given task $x$ is thus represented as an ordered sequence of calls (illustrated on the right side of Figure~\ref{fig:data_interface}), with each call comprising the component, its input, output, and the associated reward (typically only available for the final call).

Importantly, our unified data interface captures all state changes, including those caused by non-LLM components, enabling flexible and highly customizable optimization methods. For example, it supports selective optimization of specific agents within a multi-agent system (more details are in Section~\ref{sec:rl_data} and experiments in Section~\ref{sec:res_spider}), as well as optimizations beyond model finetuning, such as prompt tuning. Although this paper focuses on LLM-centric optimization, this extensibility demonstrates the broad applicability of our approach. Additional implementation details can be found in our GitHub repository\footnote{\url{https://github.com/microsoft/agent-lightning/tree/main/examples/apo}}.

By capturing the complete execution context in each state, our unified data interface provides sufficient information for policy learning, achieving a clean decoupling between agent execution and RL training. This design accommodates arbitrary and complex agent interaction logic without requiring explicit parsing of the entire execution DAG, thereby greatly simplifying RL-based optimization for diverse agent workflows.

\subsection{Markov Decision Process in Agents}
\label{sec:method_mdpa}

\subsubsection{Formulation}

Let's begin with a simple case: an agent containing a single LLM to be optimized. When we treat this LLM as a policy model, we can model its decision-making process as a  Partially Observable Markov Decision Process (POMDP), which forms the foundation for applying RL. The POMDP tuple $\mathcal{M} = (\mathcal{S}, \mathcal{O}, \mathcal{A}, \mathcal{P}, \mathcal{R})$ can be defined as:
\begin{itemize}
    \item $\mathcal{S}$ is the space of states, corresponding to all possible states, i.e., $\texttt{state}_t\in\mathcal{S}$;
    \item $\mathcal{O}$ is the observation space, corresponding to all possible inputs to the policy LLM, i.e., $\texttt{input}_t\in\mathcal{O}$;
    \item $\mathcal{A}$ is the action space, where a single action is defined as the entire token sequence generated from a single invocation of the policy LLM, i.e., $\texttt{output}_t\in\mathcal{A}$.
    \item $\mathcal{T}(s'|s,a)$ defines the (usually unknown) transition dynamics to the new state;
    \item $\mathcal{R}(s,a)$ is the reward function, which maps state-action pairs to scalar rewards;
\end{itemize}

Concretely, at each step $t$, the agent snapshot is $\texttt{state}_t$, and the LLM observes a context $\texttt{input}_t$. As explained in Section~\ref{sec:state_call}, $\texttt{input}_t$ is a semantic variable representing the part of the $\texttt{state}_t$ that is visible or necessary for the policy LLM to make decisions. The LLM then generates a sequence of tokens $\texttt{output}_t = (y_{t,1}, y_{t,2}, \dots, y_{t, N_t})$ as its output. This sequence is treated as a single action, denoted by $a_t \in \mathcal{A}$. After executing $a_t$, the agent transitions to a new $\texttt{state}_{t+1}$. A scalar reward $r_t = \mathcal{R}(s_t, a_t)$ may be provided to evaluate the quality of the action. The return is defined as the sum of rewards:
\[
R=\sum_{t=1}^T r_t.
\]
The policy model's goal is to maximize the return.

\subsubsection{Data Extraction for RL}
\label{sec:rl_data}

Based on the MDP formulation above, we need to collect trajectories containing all policy LLM decisions and their associated rewards for RL. To this end, we extract from each execution (i.e., Eq.~\eqref{eq:execution_r}) only the relevant information for updating the policy LLM parameterized by $\theta$: the raw input, output of LLM calls and their rewards, denoted as:
\begin{align}
\label{eq:execution_r_rl}
&\texttt{execution}^{RL}(x,k) = \left\{( \texttt{input}_t^{x,k}, \texttt{output}_t^{x,k},r_t^{x,k}) \right\}_{t=1}^{T},\\
&\text{with}~\texttt{output}_t^{x,k} = \pi_{\theta}(\texttt{input}_t^{x,k}).
\end{align}
Here, $\pi_\theta$ is the LLM to be optimized and $T$ is the number of invocations in this trajectory. Figure~\ref{fig:data_interface} demonstrates this data extraction process.

LLM calls are stateless and highly flexible. Their input can encode various information produced by the complex agent logic, including but not limited to instruction prompts specifying the LLM's role or goal, conversation history, and content such as user queries, reasoning steps, retrieved documents varying across tasks and executions. The agent framework may further rendered or parsed these inputs and outputs (e.g., via templates or structured formats).
For RL training purposes, thanks to the MDP formulation, we can ignore the cumbersome and variable logic and processing, focusing solely on the current input and output of the LLM.
This approach omits the detailed reasons and sources for obtaining LLM's inputs. \textit{Such extraction help us avoid tedious but non-trivial parsing of traces generated by highly dynamic agent runs, making it feasible to apply RL to optimize ANY AI agent.}

In summary, the unified data interface and MDP formulation allow   \ours{} to clearly separate  task-specific agent design from learning-based policy optimization, laying the foundation for effective RL-based optimization in modular and dynamic agent systems.

\paragraph{Application to single-LLM multi-agent setting}

Note that the above formulation flexibly applies to multi-agent scenarios with a single LLM, where the LLM can take on multiple roles at different stages of execution based on the prompt. In the RAG case in Figure~\ref{fig:data_interface}, during the first LLM call, we can instruct the model to focus on search, generating a search query $a_1$. This query modifies the $\texttt{state}$, forming new $\texttt{input}$. During the second LLM call, we can instruct the model to focus on answering the given question, generating the final answer $a_2$. The quality of this answer determines the final reward $r_2$. In this way, a single LLM can act as two agents. Moreover, \ours{} enables selectively optimizing agents within a multi-agent system, by including the corresponding agent's transitions in the optimization process. Compared to the masking approach, selecting transitions is much more convenient and intuitive.

\paragraph{Extension to multi-LLM setting}

When multiple distinct LLMs with their own parameters must be jointly optimized, a straightforward approach treats each LLM as an independent MDP and optimizes them separately. While this simplifies training, it ignores inter-dependencies between policies and may lead to suboptimal coordination. A more principled approach would use multi-agent reinforcement learning (MARL) or game theory, treating each LLM as an agent with its own objective and policy ~\citep{lowe2017multi,zhang2021multi}. Such approaches better capture potential interaction dynamics and competitive or cooperative behavior among multiple LLMs.

\subsection{\ourrlalg{}: A Hierarchical RL Method for Optimizing LLMs in Agents}
\label{sec:method_hrl}

\subsubsection{Preliminary on Single-Turn Reinforcement Learning for LLMs}

Recent advances in RL for LLMs have focused on single-call problems, where the model generates a response to a prompt in one pass. These algorithms are primarily applied in domains such as mathematics and logic puzzles~\citep{guo2025deepseek,xie2025logic}. In this setting, given a task \( x \), an LLM generates a response \( \texttt{output} = (y_1, \ldots, y_N) \) token by token from a policy \( \pi_\theta \), and each token is treated as an action. After the entire response is generated, a scalar reward \( r(x, \texttt{output}, \hat{y}) \) is assigned based on the correctness or quality of the solution, where $\hat{y}$ is the ground truth to task $x$. The typical token-level loss can be simplified as\footnote{Note that some improvements and constrains, such as importance sampling ratio, clipping, and KL divergence terms are omitted in this equation, but they are adopted accordingly in the final algorithm.}:
\begin{equation}
\mathcal{L}(\theta) = -\mathbb{E}_{x \sim \mathcal{X},\, \texttt{output} \sim \pi_\theta(\cdot|x)}\left[ \sum_{j=1}^N \log \pi_\theta(y_j | x, y_{<j}) \cdot A_j \right],
\label{eq:single-rl-objective}
\end{equation}
where \( A_j \) is a token-level advantage estimate, and $\mathcal{X}$ is the task set. The core distinction among popular algorithms lies in the estimation of the advantage term, $A_t$. Standard PPO~\citep{ouyang2022training} employs a learned critic model, defining the advantage with a value function. In contrast, recent methods eliminate the need for a separate value function to simplify training. For instance, GRPO~\citep{shao2024deepseekmath} calculates the advantage by normalizing the reward of a response against the mean and standard deviation of rewards from a group of responses generated for the same prompt. Similarly, REINFORCE++~\citep{hu2025reinforce++} uses a simpler baseline, defining the advantage as the difference between the current response's reward and the average reward of the entire training batch.

\subsubsection{Extend to Agent Scenarios via \ourrlalg{}}

In the above MDP formulation, we treat the entire sequence of tokens generated during a single LLM invocation as one action. The policy may need to take multiple such actions through interactions to complete a full episode. In contrast, LLMs generate outputs token by token, and most existing RL algorithms for LLMs are designed for single-turn interactions, focusing on optimizing token generation within a single response. To bridge this gap, \ours{} introduces a simple hierarchical reinforcement learning (HRL) approach, called \ourrlalg{}. This approach seamlessly integrates with existing single-turn RL methods for LLMs without modifications and effectively supports optimization in any agent scenario.

\begin{figure}[tbp]
    \centering
    \includegraphics[width=1.0\textwidth]{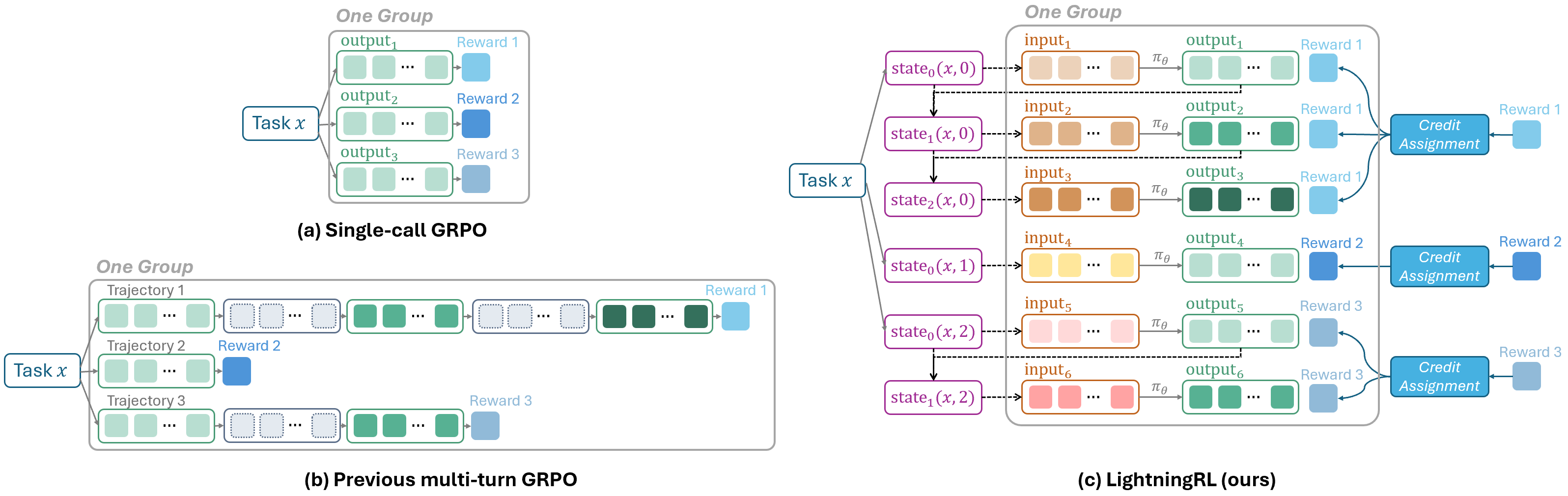}
    \caption{Illustration of the \ourrlalg{} algorithm. \textbf{(a)} Single-call GRPO, where the LLM generates a response to a task in one pass. Outputs for the same task are grouped together for advantage estimation. \textbf{(b)} Previous multi-turn GRPO. Each trajectory contains multiple LLM calls, with trajectories for the same task grouped for advantage estimation. Tokens not generated by the LLM are masked (gray dashed boxes) during optimization. \textbf{(c)} Our proposed \ourrlalg{}. Trajectories are decomposed into transitions, and transitions for the same task are grouped for advantage estimation. Each transition includes the current input/context, output, and reward. The input is part of the current agent state, with rewards computed by the credit assignment module.}
    \label{fig:alg}
\end{figure}

As shown in Figure~\ref{fig:alg}, \ours{} organizes data using transitions, and each transition is defined as the tuple $(\texttt{input}_t^{x,k}, \texttt{output}_t^{x,k}, r_t^{x,k})$ in Eq.~\eqref{eq:execution_r_rl}. Each transition contains the input context for the current LLM call, regardless of how the context is constructed. \ourrlalg{} applies a two-step mechanism: the episode-level return $R$ is first assigned across actions by a credit assignment module; then it is further decomposed across tokens within each action, producing token-level supervision signals. In our current implementation, \ourrlalg{} simply assumes each action within the episode has the same value, equal to the final return $R$. The second step is then handled by existing single-turn RL algorithms for LLMs.

This design brings several advantages. \textit{First, it allows direct use of any single-turn RL algorithm without modification, especially value-free methods that are typically lightweight and efficient.} For example, in GRPO, samples generated from the same prompt are grouped to estimate the advantage. We apply the same approach here: each task $x$ is run multiple times to generate different execution data, each decomposed into individual actions. These samples are then grouped by task for computing statistics in Eq.~\ref{eq:single-rl-objective}. Similar adaptations apply to PPO and REINFORCE++. 

\textit{Second, \ourrlalg{} enables flexible construction of policy observations}, since our data is organized at the level of individual transitions, each with its own reward. Current input/observation can be flexibly derived from the state. For example, $\texttt{input}_t$ can be a summary of previous steps generated by an LLM, or a structured prompt assembled via templates, or an instruction indicating the current role of the LLM (explained in Section~\ref{sec:rl_data}). In contrast, previous methods often concatenated all turns in a trajectory into a single response and used masking to control which parts were updated--an
 approach that cannot support such flexible and modular context construction.

\textit{Third, compared to the masking strategy, \ourrlalg{} offers a more robust and scalable implementation.} Masking-based approaches not only require tight coupling between training and agent execution logic, but also disrupt the continuity of tokens in LLMs, which is assumed in the widely used position encoding approaches, such as Rotary Positional Embeddings (RoPE)~\citep{su2024roformer}. Additionally, masking introduces significant complexity in code verification, debugging, and kernel design, often resulting in reduced efficiency when masks become intricate. In contrast, \ourrlalg{} decomposes agent trajectories into transitions, naturally aligning with the LLM’s input structure and eliminating the need for additional masking. This design simplifies implementation and enhances scalability. Moreover, organizing data in the form of transitions alleviates the issues caused by accumulative context and excessively long contexts resulting from concatenating all turns. Long trajectories are broken down into batches of transitions, enabling the use of techniques like batch accumulation for efficient updates.

\paragraph{More sophisticated credit assignment}

The credit assignment module allows for future integration of more sophisticated strategies, such as assigning credit based on heuristics or learned models. One potential avenue for future work is introducing a high-level value function to estimate the expected return for each action $a_t$ individually, providing a more nuanced credit distribution. Nevertheless, our experimental results demonstrate that the simple identical assignment strategy is effective across multiple scenarios and datasets.

\subsection{System Design of \oursnorm{}}
\label{sec:method_infra}

Training LLMs for real-world agents is a challenging endeavor that requires the integration of RL training frameworks and agent development frameworks. These frameworks are often complex, rapidly evolving, and inherently fragmented, making them difficult to align. The \ours{} Framework addresses this challenge by offering a unified solution that facilitates the seamless development and training of agents using RL and other optimization techniques.

\begin{figure}[tbp]
    \centering
    \includegraphics[width=.6\textwidth]{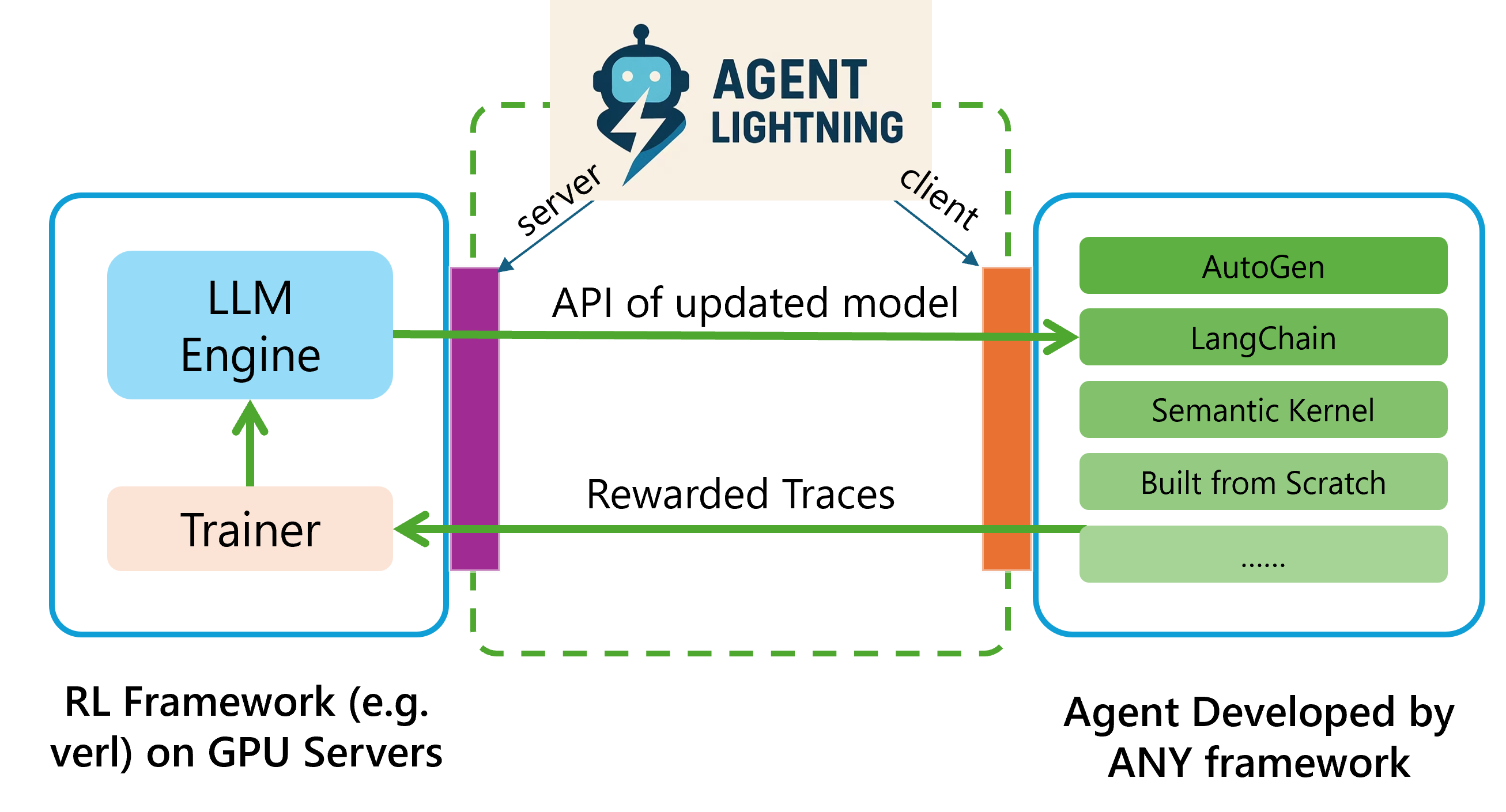}
    \caption{Training-Agent Disaggregation architecture.}
    \label{fig:agent_lightning_architecture}
\end{figure}

\subsubsection{Training-Agent Disaggregation Architecture}

The RL training framework for LLMs consists of two primary components: the trainer and the rollout. The trainer is responsible for updating the model weights, while the rollout captures trajectories, which are data used by the trainer.
For single-turn RL tasks, such as mathematical reasoning, the rollout process is straightforward, involving only the generation of LLM responses given specific prompts. Conversely, for agentic tasks that involve multiple turns and interactions with environments, the rollout necessitates the complete execution of agents. This includes not only multiple instances of LLM generation but also the execution of native code, such as tool utilization.

The core innovation of \ours{} is its {Training-Agent Disaggregation} architecture, which decouples the compute-intensive LLM generation and light-weighted but diverse and flexible application logic and tools written in traditional programming language. The former is managed by the RL framework, and expose an OpenAI-like API to the latter, illustrated by Figure~\ref{fig:agent_lightning_architecture}. The latter can be managed and executed independently, no need to be collocated with the GPU resource, allowing for greater flexibility in defining agentic behaviors.

\paragraph{Agent-Agnostic Training and Trainer-Agnostic Agents}
A key advantage of the \ours{} Training-Agent Disaggregation architecture is the mutual independence it establishes between the training framework and the agent. This design renders the training framework (e.g., VeRL~\citep{sheng2024hybridflow}) \textit{agent-agnostic}; its sole concern is the optimization of the LLM and management of hardware resources, without being coupled to specific agent logic. Conversely, the agent, operating on the client side, is \textit{trainer-agnostic}, allowing it to function independently of the training framework's implementation details. This decoupling provides significant flexibility in agent development and deployment, empowering developers to concentrate on agentic logic without the constraints of the training infrastructure. Appendix~\ref{sec:appendix_code_example} provides a code example demonstrating how an existing agent can be readily optimized using \ours{}, showcasing its ease of use and adaptability.

To fulfill above design goals, \ours{} introduces a two-component architecture: the \ours{} Server and the \ours{} Client, as depicted in Figure~\ref{fig:agent_lightning_architecture}. The server is designed to integrate seamlessly with RL frameworks, managing the training process and LLM optimization. It handles the orchestration of tasks, data management, and communication with the client. The client encapsulates agents, allowing them to operate independently of the training framework, thus enabling flexibility in agent development and execution.
Upon receiving the task dataset uploaded by the user, the server, initialized together with the RL framework, orchestrates the training procedure. The server divides the dataset into batches and manages the transitions between generation and training stages using an event-driven system, in coordination with the client. Task batches are dispatched to clients via an inventory manner, where the server maintains a record of available tasks and assigns them to clients as they become ready. An OpenAI-like API endpoint, which is unique for each task, is passed to client simultaneously with the task. This endpoint allows the client to connect to the server and facilitates the data tracing and capture process.
The client then executes the agents in agent runtime (Section~\ref{sec:agent_runtime}) to generate traces and rewards. The traces are collected and reported back to the server, which processes the data and forwards it to the training framework for model parameter updates. This architecture allows for efficient and scalable training of agents, leveraging the strengths of both the RL framework and the agent development framework.

\subsubsection{Agent Runtime}
\label{sec:agent_runtime}
The \ours{} client is a runtime manager for agents that orchestrates execution, captures data, handles errors, and communicates with the server. It supports the agent training workflow to be robust, scalable, and efficient.

\paragraph{Data Parallelism for Agent Execution}
Large batch sizes are essential in modern RL training to fully utilize computational resources and reduce rollout latency, which are the bottleneck of the training loop. To process these large batches, multiple agent instances must run in parallel.  The \ours{} client is designed to efficiently manage and execute multiple agent instances concurrently, leveraging data parallelism to maximize throughput and minimize latency. Particularly, it implements a two-level parallelism strategy: intra-node parallelism, where one client runs multiple workers on a single machine, each executing a different agent instance, and inter-node parallelism, where multiple clients run on different machines, each managing its own set of agent instances. This architecture allows for flexible scaling and efficient resource utilization across distributed systems.

\paragraph{Data Capture without Code Modification}
To integrate seamlessly with existing agent codebases, the Lightning client employs two instrumentation techniques that capture relevant data without requiring agent logic modifications. The first technique is based on OpenTelemetry~\citep{OpenTelemetry} and AgentOps~\citep{AgentOps}. It uses OpenTelemetry's tracing capabilities to automatically instrument agent code, capturing execution traces, LLM calls, and environment interactions. For users who prefer not to rely on OpenTelemetry, \ours{} also provides a basic tracing mechanism embedded in the OpenAI-like API endpoint. This mechanism is lightweight and scalable, ensuring that data capture can be performed efficiently even for agent development frameworks that are not compatible with OpenTelemetry.

\paragraph{Error Handling and Robustness}
The stability and reliability of the training process are paramount and sensitive to various types of failures, especially in long-running training sessions. The errors can arise from various sources, including agent crashes, network interruptions, or invalid outputs. And their frequency can significantly increase in RL training than in inference due to RL algorithms' inherent exploration and exploitation dynamics. Such errors, if not handled properly, can lead to significant downtime and wasted computational resources. To address this, the \ours{} client incorporates comprehensive error handling mechanisms that ensure robustness during training. It detects failures that are not properly handled by the agent code, such as crashes or long-hanging tool calls, and ensures that these failures do not disrupt the overall training process. Failed tasks can be retried or reassigned to other agent instances, and detailed logs are maintained for debugging and monitoring purposes. This design minimizes downtime and maximizes training efficiency, allowing for continuous improvement of agent performance.

\paragraph{Automatic Intermediate Rewarding (AIR)}
The delayed and sparse nature of rewards in RL training can hinder the learning process. Intermediate rewards, which provide feedback at various stages of agent execution, can significantly enhance the training process by offering more frequent and informative signals. But it often requires high overhead such as costly human annotation or complex reward computation logic. How to mine intermediate rewards from the agent execution is of great importance to the training process. The \ours{} client provides AIR mechanisms to convert system monitoring data into intermediate rewards, enabling the training framework to leverage these signals for more effective learning.

\paragraph{Environment and Reward Services for Scalability}
Environments and reward functions are critical components of RL training. For those environments and reward functions that are light-weight and can run locally, they can be executed directly within the same worker with the agent instance. But for those that are resource-intensive or with higher initialization costs, such as mobile phone emulators or complex reward computation logic, it is more efficient to host them as shared services. Currently, the \ours{} client supports hosting services for environments and reward computation in a simple pooling manner, and it can be extended to support more complex serverless architectures in the future.

\section{Results}

The effectiveness of \ours{} is validated on three distinct tasks, each implemented with a different agent framework. The summary of the example tasks are in Table~\ref{tab:exp_tasks}. As shown by results in the following subsections, \ours{} enables continuous and stable performance improvement across diverse scenarios, demonstrating strong potential for real-world agent optimization.

\begin{table}[ht]
\centering
\caption{Summary of the tasks and settings in experiments.}
\label{tab:exp_tasks}
\begin{tabular}{lccc}
\toprule
\textbf{Task} & \textbf{Text-to-SQL} & \textbf{Open domain QA} & \textbf{Math QA}  \\
\midrule
Framework & LangChain & OpenAI Agents SDK & AutoGen \\
Dataset & Spider & MuSiQue & Calc-X \\
Tool used & SQL executor & Wikipedia retriever & Calculator \\
Num. of agents & 3 & 1 & 1 \\
Num. of tuned agents & 2 & 1 & 1 \\
\bottomrule
\end{tabular}
\end{table}

\subsection{Text-to-SQL via LangChain}
\label{sec:res_spider}

In this task, given a natural language question and a database, the agent must generate a SQL query to retrieve relevant information, and then answer the question. We use the Spider dataset~\citep{yu2018spider}, a complex and cross-domain text-to-SQL benchmark designed to evaluate a model’s ability to generate executable SQL queries from natural language questions. It contains over 10,000 questions across 200 databases with diverse schemas covering 138 different domains, requiring models to generalize to unseen databases at test time. We use $\texttt{Llama-3.2-3B-Instruct}$~\citep{meta2024llama} as the base model.

This case is implemented using LangChain, and designed as a multi-agent system. It contains 3 agents and the workflow is as follows. The SQL writer agent first generates a SQL query, which is then executed. If the query is correct, the SQL executor returns information from the database; if incorrect, it returns error messages. Next, a checking agent evaluates the correctness of the SQL query and the validity and sufficiency of the retrieved information. This agent decides whether to rewrite the query or directly generate an answer. If rewriting is needed, the re-writing agent revises the query based on the checker LLM's feedback. If no rewriting is required, the re-writing agent also being responsible for question-answering, takes the retrieved information and the question to generate the final answer. In this workflow, SQL generation, checking, and re-writing are performed by the same LLM, but with different prompts tailored to each task, effectively functioning as 3 agents. And during training, we only optimize two of them, i,e, the SQL writing and re-writing agents. And these two agents are optimized simultaneously. This case demonstrates that \ours{} can selectively optimize one or more agents within a multi-agent system.

This task requires the LLM to possess a range of capabilities, including generating and correcting syntactically and semantically valid SQL queries, checking the quality of the query and retrieved information, understanding the current context to assess whether sufficient information has been retrieved, improving the SQL to obtain the correct information, and finally answering the question based on the retrieved results.

The reward for this task is determined by whether the final answer to the question is correct. The model’s final performance is also evaluated based on its answer accuracy on the test set.

\begin{figure}[ht]
    \centering
    \begin{subfigure}{0.48\linewidth}
        \centering
        \includegraphics[width=\linewidth]{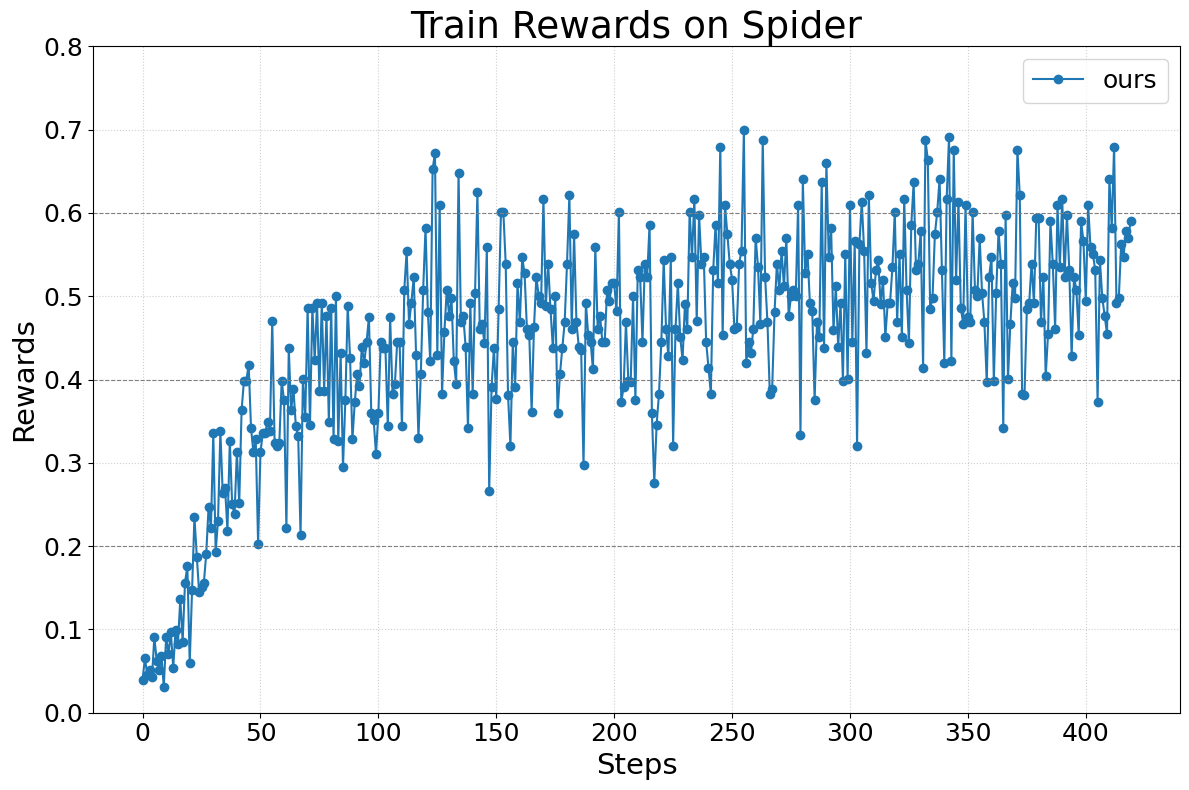}
        \caption{Train reward}
    \end{subfigure}
    \hfill
    \begin{subfigure}{0.48\linewidth}
        \centering
        \includegraphics[width=\linewidth]{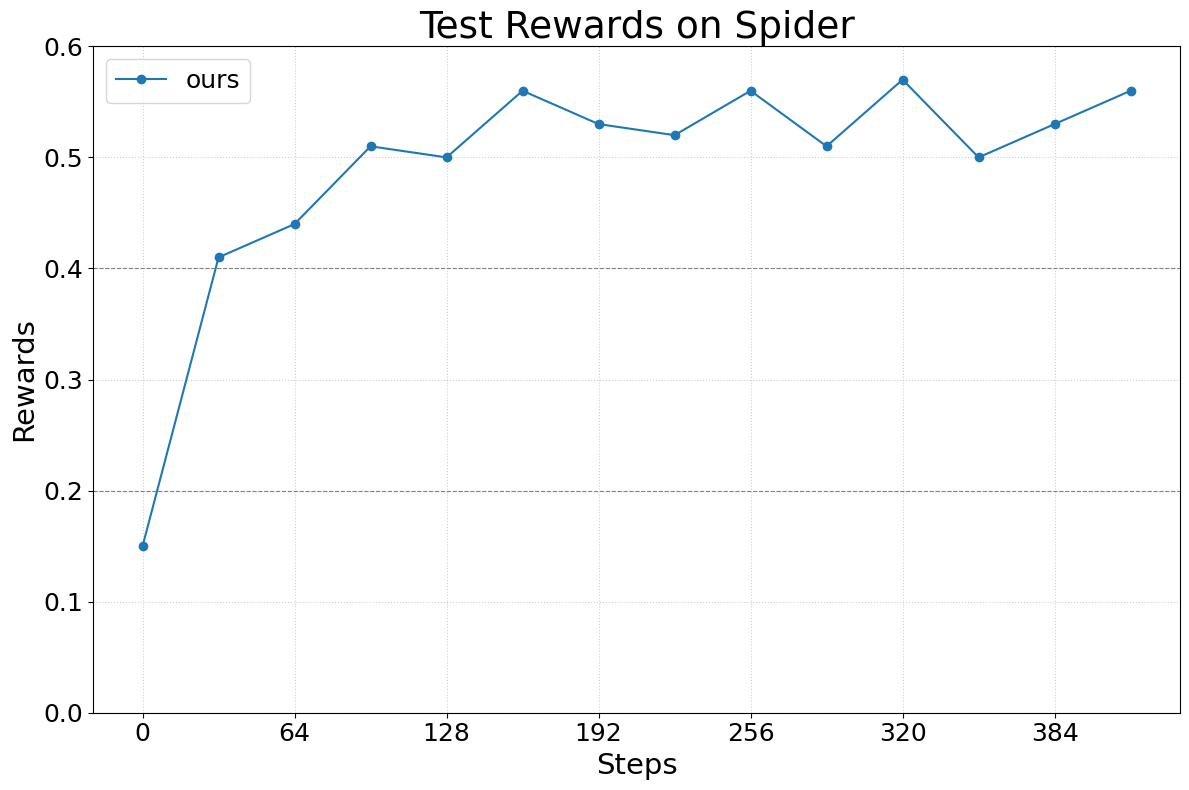}
        \caption{Test reward}
    \end{subfigure}
    \caption{Reward curves for the Text-to-SQL task.}
    \label{fig:sql_curve}
\end{figure}

As shown in Figure~\ref{fig:sql_curve}, \ours{} enables stable reward improvement, demonstrating its ability to optimize complex multi-step decisions involving code generation and tool use.

\subsection{Retrieval-Augmented Generation via OpenAI Agents SDK}
\label{sec:res_rag}

The second one is a typical Retrieval-Augmented Generation (RAG) task. Given a question and a document database, the agent first generates a natural language query to retrieve supporting documents using an existing retriever tool. We use MuSiQue dataset~\citep{trivedi2021musique}, a challenging multi-hop question answering benchmark designed to promote genuine compositional reasoning. Unlike earlier benchmarks that can often be solved using shallow shortcuts, MuSiQue is constructed using a bottom-up methodology that systematically composes connected single-hop questions, ensuring that one reasoning step depends critically on the output of another. Besides, to mimic real-world search setting, the given database is entire Wikipedia, containing 21 million documents. And we use embeddings generated by BGE model~\citep{bge-large-en-v1.5,zhang2023retrieve} and cosine similarity as our retriever tool. These multi-hop questions and the large searching source exhibit great challenge for agents. We use $\texttt{Llama-3.2-3B-Instruct}$~\citep{meta2024llama} as the base model.

This agent is implemented using the OpenAI Agents SDK. The agent workflow here is similar with but simpler than the previous text-to-SQL task. The policy LLM needs to first generate a query, and then decides whether to refine the query or generate an answer according to retrieved documents. One LLM is employed in this workflow and we do not use the multi-agent setting in this case, meaning that only one instruction containing all things that needs to be done by the LLM is used. Unlike the SQL task, the queries here are free texts, making the retrieval and reasoning steps more open-ended. Also, the given database is much larger than the previous task. This task tests the agent’s capability to formulate effective retrieval queries with semantic meanings and reason over retrieved textual information. 

The reward employed during training is a weighted combination of a format score $R_{\text{format}}$ and a correctness score. Specifically, the format score is 1 if the LLM can output in a specific format, where the corresponding content must be enclosed by specific tokens, such as $\texttt{<think>...</think>}$, $\texttt{<query>...</query>}$ and $\texttt{<answer>...</answer>}$. And the correctness score $R_{\text{correctness}}$ is computed as the word-level F1 score between the predicted answer and the gold answer. Two scores are combined as $R=0.9 \times R_{\text{correctness}} + 0.1  \times R_{\text{format}}$.

\begin{figure}[ht]
    \centering
    \begin{subfigure}{0.48\linewidth}
        \centering
        \includegraphics[width=\linewidth]{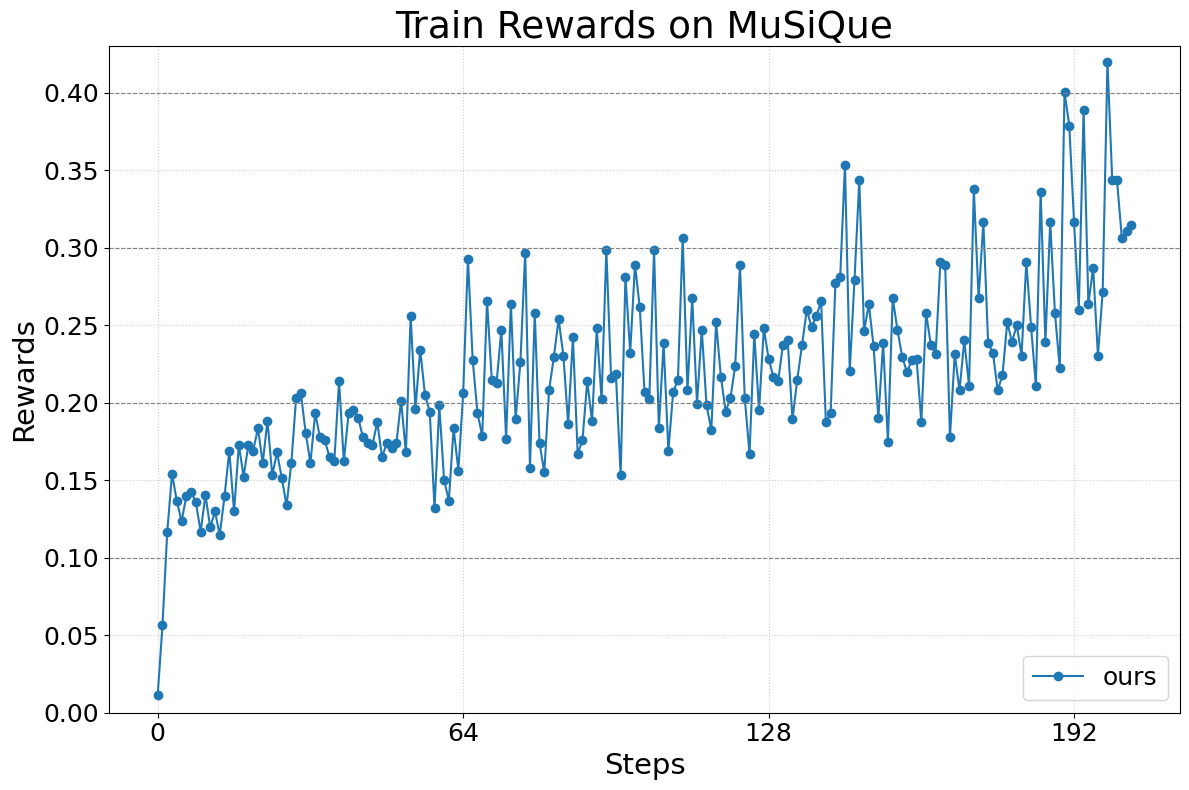}
        \caption{Train reward}
    \end{subfigure}
    \hfill
    \begin{subfigure}{0.48\linewidth}
        \centering
        \includegraphics[width=\linewidth]{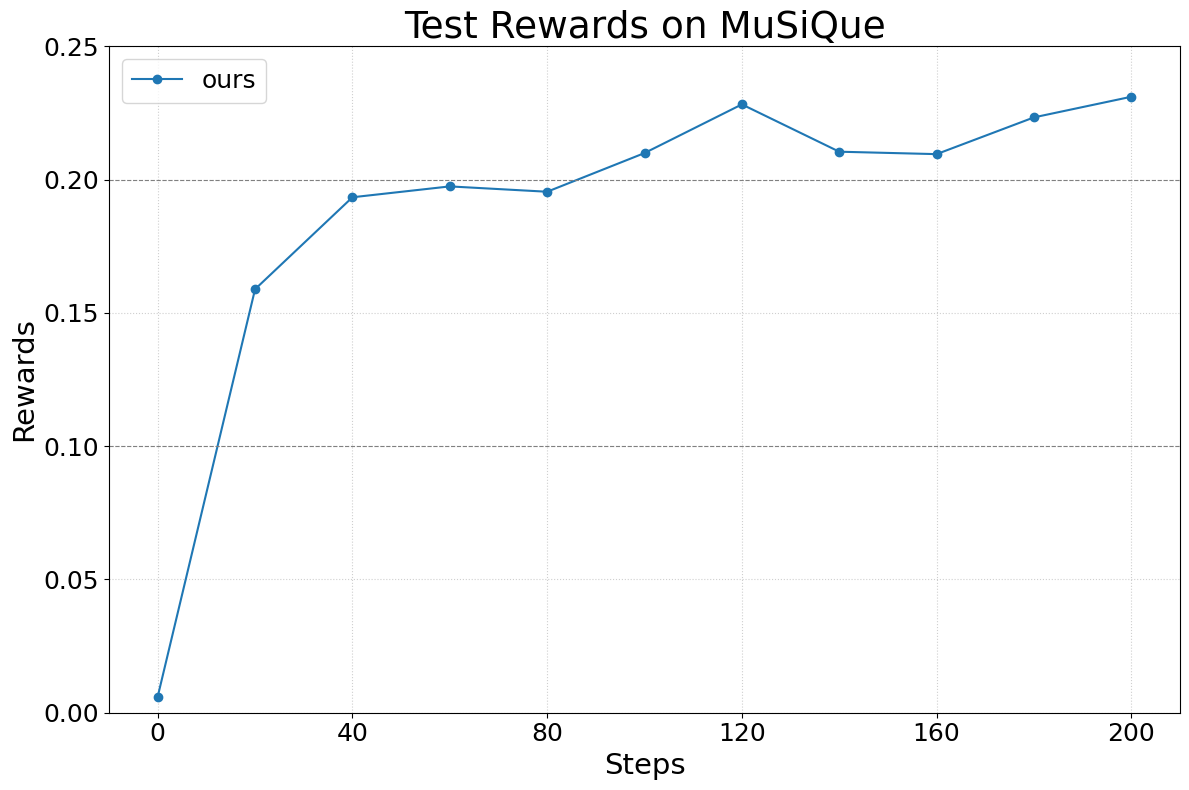}
        \caption{Test reward}
    \end{subfigure}
    \caption{Reward curves for the Retrieval-Augmented Generation task.}
    \label{fig:musique_curve}
\end{figure}

Figure~\ref{fig:musique_curve} demonstrates that \ours{} enables stable performance improvement on this challenging task, demonstrating its effectiveness in more complex and open-ended RAG scenarios.

\subsection{Math QA with Tool Usage via AutoGen}
\label{sec:res_math}

The third task is a math-focused QA task that evaluates the agent’s ability to invoke tools (specifically, a calculator) to solve arithmetic and symbolic problems. We use the Calc-X dataset~\citep{kadlcik-etal-2023-calcx}, which consists of diverse math problems requiring reasoning and precise computation. By modifying existing math datasets, such as GSM8K and Ape210K, Calc-X emphasizes the integration of external tools into reasoning workflows, making it well-suited to test tool-augmented agents. We use $\texttt{Llama-3.2-3B-Instruct}$~\citep{meta2024llama} as the base model.

The agent is implemented via AutoGen and follows a simple yet tool-intensive workflow. Given a natural language math question, the agent must decide how and when to invoke a calculator tool to compute intermediate values before producing the final answer. The workflow is executed by a single LLM, which is responsible for generating tool calls, interpreting tool outputs, and forming the final answer. This requires the model to understand math problem structures, issue syntactically correct tool calls, and properly integrate tool outputs into the final reasoning chain.

The final reward is based on whether the agent correctly answers the question, and the model's performance is also evaluated via answer accuracy on the test set.

\begin{figure}[ht]
    \centering
    \begin{subfigure}{0.48\linewidth}
        \centering
        \includegraphics[width=\linewidth]{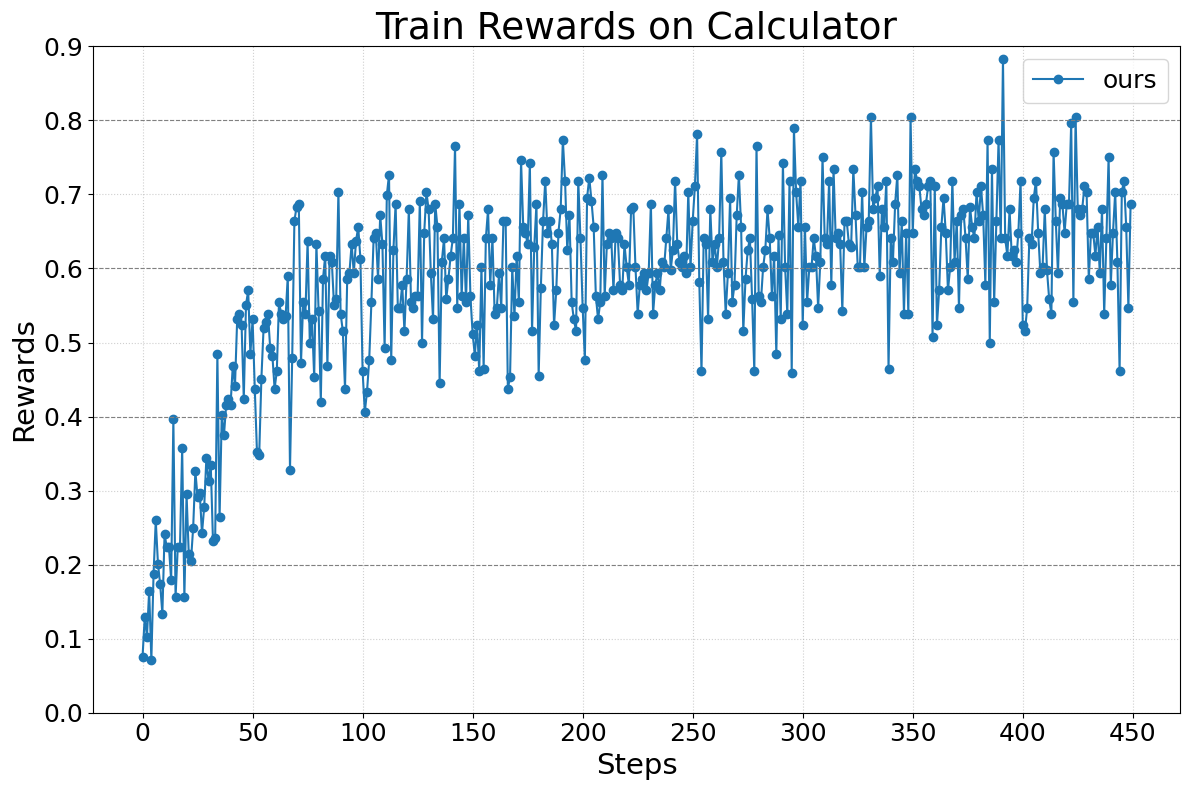}
        \caption{Train reward}
    \end{subfigure}
    \hfill
    \begin{subfigure}{0.48\linewidth}
        \centering
        \includegraphics[width=\linewidth]{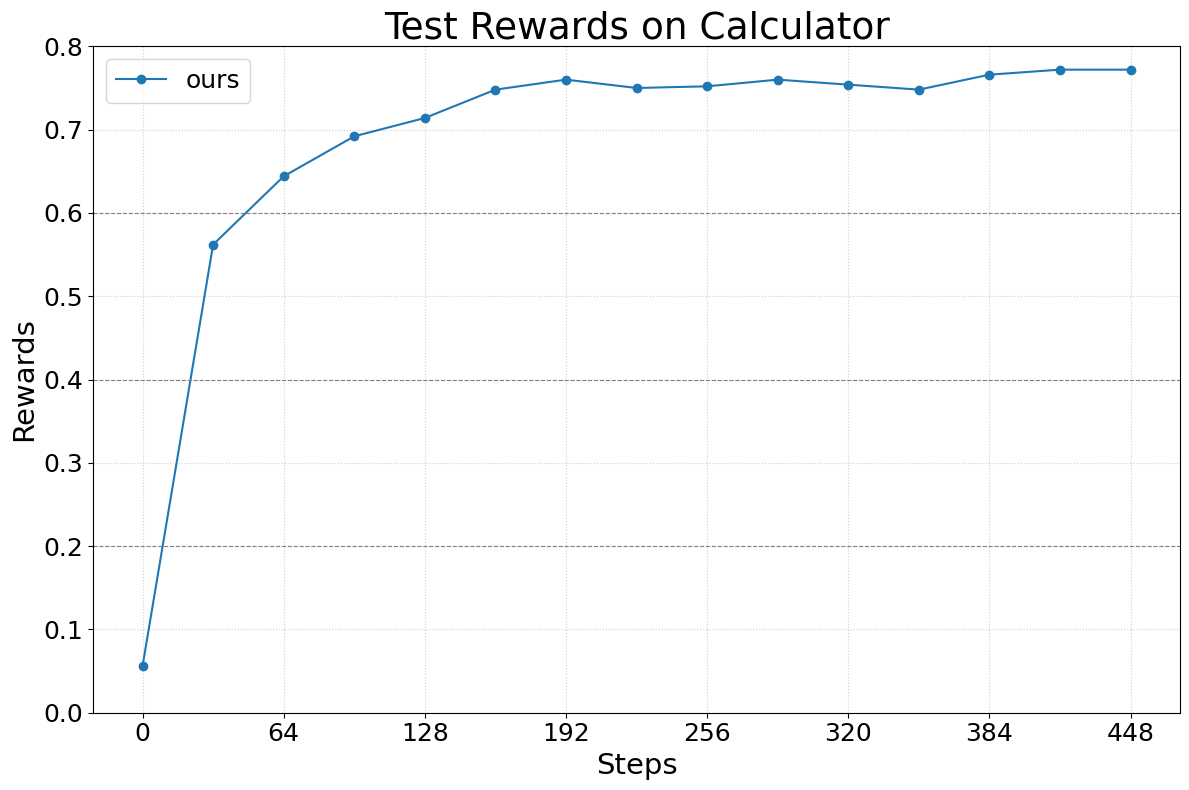}
        \caption{Test reward}
    \end{subfigure}
    \caption{Reward curves for the Calculator task.}
    \label{fig:calc_curve}
\end{figure}

As shown in Figure~\ref{fig:calc_curve}, \ours{} improves performance consistently over training. This demonstrates its effectiveness in tool-augmented settings, where precise external function calls and reasoning are both required.

\section{Discussion}

\subsection{Related Work}

\paragraph{Recent Advances in Multi-Turn RL for LLMs}

With the advancement of RL for LLMs and the emergence of open-source RL training frameworks such as verl~\citep{sheng2024hybridflow}, there is increasing interest in extending these frameworks to support more sophisticated agent training scenarios. Recent works, including RAGEN~\citep{wang2025ragen}, Trinity-RFT~\citep{pan2025trinity}, rLLM~\citep{rllm2025}, Search-R1~\citep{jin2025search,THUDMslime} and etc., have explored RL in multi-turn, interactive environments. These methods typically concatenate agent turns into a single, elongated sequence and assigning specific masks to ensure correct optimization.

Instead of concatenating, we leverage our MDP formulation, unified data interface, and credit assignment module to organize data into individual transitions, which offers several benefits in capturing the complex interaction logic characteristic of real-world agents. \textit{First}, our transition-based modeling supports a wide range of agent architectures and workflows~\citep{HowWeBuilt2025}, including multi-agent orchestration. This flexibility bridges the gap between training and deployment, ensuring that RL-based optimization aligns with the dynamic and diverse nature of real-world agents. Concatenation-based methods are only suitable for a narrow subset of agents with simple, sequential workflows and struggle to handle more sophisticated patterns.
\textit{Second}, by organizing data as transitions rather than concatenated turns, our approach relieves the issue of accumulative context length. As agents accumulate context through multi-turn interactions, tool outputs, protocol exchanges (e.g., MCP~\citep{mcp}), and multi-agent communication, the resulting sequences frequently exceed the input length limitation of LLMs or raise higher computation and memory requirement for the training services. Viewing a transition as a single sample significantly alleviates the sequence length increase caused by turn accumulation.
\textit{Third}, our framework eliminates the need for custom masking strategies since only current LLM input is included in the training sample. Whereas, to ensure correct optimization, concatenating necessitates the design of custom masking strategies for inputs, losses, and attention mechanisms. These masking strategies are often application-specific, non-generalizable, and require tight coupling between the training process and agent execution logic. They also disrupt positional continuity, making implementation and debugging more difficult.
\textit{Fourth}, the transition-based formulation unlocks the potential for advanced RL algorithms, such as hierarchical RL algorithms~\citep{zhou2024archer}, which offer more effective credit assignment mechanisms. By enabling algorithmic flexibility, our approach paves the way for innovative RL methods tailored to complex agent scenarios.

\paragraph{Large-Scale RL Training Systems for LLMs}

The development of RL for LLMs has been driven by stable and robust training systems, including verl~\citep{sheng2024hybridflow}, OpenRLHF~\citep{hu2024openrlhf}, TRL~\citep{vonwerra2022trl}, ROLL~\citep{wang2025reinforcement}, AReaL~\citep{fu2025areal}, and so on. These systems primarily focus on efficient LLM training in single-turn scenarios. While they also offer extensions for agent-based applications~\citep{cao2025skyrl,verlagentloop}, they generally require developers to rebuild agents inside the training system. This requirement may arise from the fact that existing RL frameworks typically demand the training side to be aware of the agent's execution logic in order to correctly organize training data, such as determining the concatenation order and where to apply masks. As a result, agents should be implemented in ways that are tightly coupled with the training framework. However, this constraint is hard to meet given the diversity of existing agent development ecosystems. In practice, real-world agents are highly customizable and are often developed using a variety of frameworks, such as OpenAI Agents SDK~\citep{oaiagentsdk}, LangChain~\citep{LangChain}, and AutoGen~\citep{AutoGen}, or even built directly from scratch with LLM SDKs. Migrating these agents to existing RL frameworks typically requires users to manually adapt or re-implement agent execution logic to conform to the framework’s requirements, which is a process that is labor-intensive, error-prone, and difficult to scale across heterogeneous agent ecosystems. Moreover, this migration often necessitates specialized expertise in RL training frameworks, such as being familiar with Ray~\citep{ray} or other distributed systems, which can present a significant barrier for many developers. Additionally, the tight coupling imposes extra burdens on the RL training process itself, as agents may involve complex dependencies, including various MCP servers, external tools, and APIs. Integrating these components within the RL training framework introduces complexity and additional overhead, detracting from the core RL objective and complicating maintenance and scalability.

Our framework differs significantly from these works in system design, as we completely decouple RL training from agents. This eliminates the need to write data collection logic within the RL training engine and introduces almost no code modifications to the agent side. As a result, we can seamlessly integrate with various agents, whether they are developed using different frameworks or built from scratch by users. Additionally, our modeling theoretically supports integration with a wide range of RL training frameworks, further enhancing flexibility and scalability.

\paragraph{Algorithm-Centric Multi-Turn RL Works}
Multi-turn RL has been a topic of interest in RL algorithm community, with works like ArCher~\citep{zhou2024archer} exploring text games and WebShop~\citep{yao2022webshop} focusing on e-commerce tasks. These works typically involve RL agents interacting with environments over multiple turns, but they often only use relative small policy models (e.g., parameter size smaller than 1B) or Parameter Efficient Fine Tuning (PEFT) in the experiments, which makes it's hard to reuse their implementation in large-scale models.

\paragraph{Application-Specific RL Training}
Since the success of RL methods in DeepSeek-R1~\citep{guo2025deepseek}, there has been a surge of interest in applying RL in some specific scenarios, such as RAG and coding.
Search-R1~\citep{jin2025search} and R1-Searcher~\citep{songR1SearcherIncentivizingSearch2025} apply RL to train LLMs to generate better queries and improve search capabilities, similar to our RAG example in Section~\ref{sec:res_rag}. DeepSWE~\citep{deepswe2025} uses multi-turn interactions to teach LLMs how to write code, call functions, and use APIs. In our text-to-SQL example in Section~\ref{sec:res_spider}, the LLM generated simple SQL queries.
Meanwhile, some works like ReTool~\citep{feng2025retool} and SimpleTIR~\citep{xue2025simpletir} aim to enhance tool-integrated long-form reasoning with RL learning.

These works are mainly focused on specific tasks or scenarios, and they often rely on a predefined workflow in both training and inference. Thus they're not targeting at general-purpose agent training. In contrast, our approach is not limited to specific tasks or scenarios; instead, we provide a unified method for any AI agent, bridging AI agents and RL training methods.

\subsection{Future Work}

\paragraph{More Optimization Methods}

Besides RL, our modeling framework well supports other optimization methods as well, such as automatic prompt optimization mentioned in Section~\ref{sec:example_agent_execution}. More generally speaking, focusing on the key components and its invocations is a principle way for agent optimization, not limiting to RL-based methods. Therefore, we introduce the concept of \textit{Component of Interest} (CoI) to specify the subset of components in the execution trace that are subject to optimization.
For example, prompt template rendering can be treated as a tool call, and by treating this tool as a CoI, \ours{} can facilitate prompt optimization methods. This unified and extensible data structure supports comprehensive downstream optimization and analysis of agent behaviors. Extending support to more optimization methods is part of our future development roadmap.

\paragraph{Improvement on RL Algorithms}

Developing more efficient RL algorithms is key to solving model optimization in more complex agent scenarios. These aspects include but are not limited to long-horizon credit assignment, exploration algorithms, off-policy algorithms, and so on. Agent Lightning models and organizes data in terms of transitions, making it more convenient to integrate additional algorithms.

\paragraph{Advancements in RL System Infrastructure for Agents}
The infrastructure supporting RL for LLMs continues to evolve, presenting significant opportunities for co-development with agentic RL frameworks like \ours{}. One promising direction is the further disaggregation of system components—specifically, separating the trainer, rollout (inference engine), and agent workflows—to address the rollout bottleneck and enhance scalability in large-scale RL training. Exploring such architectural improvements can lead to more efficient and flexible RL pipelines. Additionally, optimizing for long-horizon tasks will benefit from coordinated innovations in both RL algorithms and system design, enabling more effective training of complex agents.

\paragraph{Efficient Serving}

For LLM serving, advanced approaches with more LLM-friendly abstraction (e.g., Parrot~\citep{lin2024parrot}) can be explored to optimize resource utilization and response times. For long-context acceleration, techniques such as sparse computing for long-context processing (e.g., Minference~\citep{jiang2024minference}) can significantly enhance performance. Additionally, better resource scheduling for serving environments and tools can further streamline operations and improve scalability in diverse deployment scenarios.

\newpage

\bibliography{main}
\bibliographystyle{colm2024_conference}

\newpage

\appendix

\newpage
\appendix
\section{Example to Optimize an Existing Agent with \oursnorm{}}
\label{sec:appendix_code_example}

Here is an example of how to optimize an existing agent using Agent Lightning. Given a simple agent that interacts with a game environment (for example, 20 questions or guess number), we can use the Agent Lightning framework to optimize its performance without modifying the agent code itself. Here is the typical folder structure for the agent:

\begin{listing}[!ht]
\begin{minted}[
frame=lines,
fontsize=\scriptsize,
]{text}
agent/
 |- data/train.jsonl
 |- environments/game_server.py
 |- agent.py
\end{minted}
\caption{Agent Code Folder Structure}
\label{lst:agent_code_folder_structure}
\end{listing}

In this example, the \code{agent.py} file contains the agent's logic function \code{agent\_function}, which receives the the endpoint of LLM and game environment, and return the answer to the game. The game server is created by \code{environments/game\_server.py} with the \code{(game\_seed, ground\_truth)} pair saved in \code{data/train.jsonl}. To optimize this agent using Agent Lightning, we can create a \code{train.py} file that uses the Agent Lightning framework to optimize the agent's performance without modifying the agent code. Here is an example of how to do this\footnote{The open-source API may undergo modifications. Please refer to the GitHub repository for the most up-to-date information.}:

\begin{listing}[!ht]

\begin{minted}[
frame=lines,
fontsize=\scriptsize,
linenos
]{python}
# train.py
from agent import agent_function
from environments.game_server import GameServer
from agent_lightning import Client, Resource, Task

# TODO: environments can be managed separately
game_server = GameServer()

# Task corresponds to the data object stored in the `.jsonl` file
# Resource is the meta data pulled from the server,
#   and containing the LLM endpoint
def agent_run(resource: Resource, task: Task):
    game_url = game_server.create_game(task.game_seed)
    answer = agent_function(resource.model_api, game_url)
    reward = game_url.score(answer, task.ground_truth)
    return reward

# start AgentLightningServer and pass its URL in an environmental variable
client = Client(os.environ["AgentLightningServerUrl"])
# you may also specify the pre-uploaded data id
client.upload_data("data/train.jsonl", test_file=None)
client.train(agent_run, nworkers=-1)
\end{minted}
\caption{Agent Lightning Training Script}
\label{lst:agent_lightning_training_script}
\end{listing}

\newpage
\section{Process Diagram of \oursnorm{}}
\label{appendix:process}

\begin{figure}[h]
    \centering
    \includegraphics[width=0.7\linewidth]{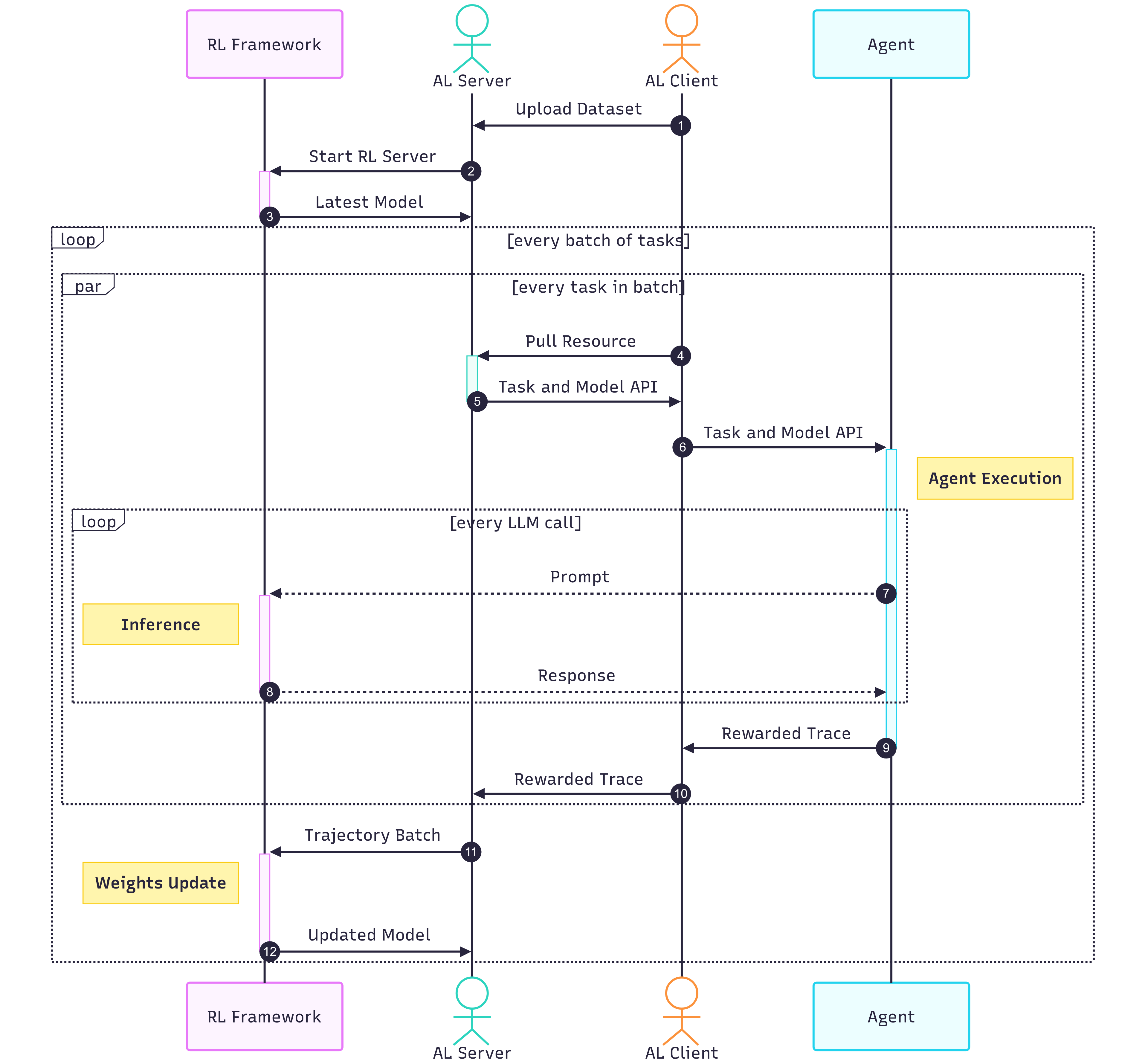}
    \caption{Process diagram}
    \label{fig:process}
\end{figure}


\end{document}